\newif\ifarxiv
\arxivtrue

\ifarxiv
\documentclass[11pt, a4]{elsarticle}
\else
\documentclass[11pt, letterpaper, twocolumn]{elsarticle}
\fi
\biboptions{sort&compress,numbers}

\usepackage[utf8]{inputenc}

\ifarxiv
    \usepackage[margin=1in]{geometry}
\else
    \usepackage[margin=1.5cm]{geometry}
\fi

\usepackage{calc}

\usepackage{tabularx}
\usepackage{lmodern}

\usepackage{placeins}

\usepackage{amsmath, amsfonts, amssymb}
\usepackage{bm}
\usepackage{booktabs}

\usepackage{hyperref}
\usepackage{cleveref}
\hypersetup{
    colorlinks=true,
    linkcolor=blue,
    filecolor=magenta,      
    urlcolor=blue,
}
\usepackage[noindentafter]{titlesec}
\titleformat{\section}[block]{\bfseries}{\thesection.}{0.5em}{} 
\titleformat{\subsection}[block]{}{\thesubsection}{0.5em}{}
\titleformat{\subsubsection}[runin]{\itshape}{}{0em}{}[\mbox{ }]

\usepackage[frozencache,cachedir=mintedcache]{minted}
\usepackage{xspace}

\ifarxiv
\usepackage[disable]{endfloat}
\else
\usepackage[nolists, nomarkers]{endfloat}  
\fi

\def\Latte{\textit{Latte}\xspace}
\def\publishedver{v0.1.0}

\newcommand{\argmax}{\operatornamewithlimits{argmax}}

\journal{Software Impacts}

\begin{document}

\ifarxiv
\else
\twocolumn[  
    \begin{@twocolumnfalse}
\fi

\ifarxiv
\begin{center}
{\Large\textbf{Latte: Cross-framework Python Package\\for Evaluation of Latent-Based Generative Models}}
\else
\noindent\textbf{Latte: Cross-framework Python Package for Evaluation of Latent-Based Generative Models}
\fi
\vskip0.5cm
\noindent
\textbf{%
    Karn N. Watcharasupat$^{1,2}$,
    Junyoung Lee$^{2}$,
    and Alexander Lerch$^{1}$%
}\\{%
    $^1$Center for Music Technology, Georgia Institute of Technology, Atlanta, GA, USA\\
    $^2$School of Electrical and Electronic Engineering, Nanyang Technological University, Singapore\\
}{Email: 
    kwatcharasupat@gatech.edu, 
    junyoung002@e.ntu.edu.sg,
    alexander.lerch@gatech.edu
}
\ifarxiv
\end{center}
\fi

\noindent
\section*{Abstract}
\noindent \Latte (for \textit{LATent Tensor Evaluation}) is a Python library for evaluation of latent-based generative models in the fields of disentanglement learning and controllable generation. \Latte is compatible with both PyTorch and TensorFlow/Keras, and provides both functional and modular APIs that can be easily extended to support other deep learning frameworks. Using NumPy-based and framework-agnostic implementation, \Latte ensures reproducible, consistent, and deterministic metric calculations regardless of the deep learning framework of choice. 

\noindent
\section*{Keywords}
\noindent Deep generative networks, disentanglement learning, latent space, controllable generation, Python

\noindent
\section*{Code metadata}
\ifarxiv
\begin{tabularx}{\textwidth}{l>{\raggedright\arraybackslash}p{5.5cm}>{\raggedright\arraybackslash}X}
    \toprule
    \textbf{Nr.} & \textbf{Code metadata description} &\\
    \midrule
    C1 & Current code version & \publishedver \\
    \midrule
    C2 & Permanent link to code/repository used for this code version & \url{https://github.com/karnwatcharasupat/latte}\\
    \midrule
    C3  & Permanent link to Reproducible Capsule & \url{https://codeocean.com/capsule/3186531/tree}\\
    \midrule
    C4 & Legal Code License   & MIT License \\
    \midrule
    C5 & Code versioning system used & Git\\
    \midrule
    C6 & Software code languages, tools, and services used & 
    Language: Python 3                  \newline
    CI/CD: pytest, CircleCI, CodeCov, CodeFactor            \\
    \midrule
    C7 & Compilation requirements, operating environments \& dependencies & 
    Python $\ge$ 3.7, $<$ 3.10 \linebreak
    NumPy $\ge$ 1.18, Scikit-learn $\ge$ 1.0.0 \linebreak
    (optional)
    PyTorch $\ge$ 1.3.1, TorchMetrics $\ge$ 0.2.0 \linebreak
    (optional)
    TensorFlow $\ge$ 2.0 
    \\
    \midrule
    C8 & If available Link to developer documentation/manual & \url{https://latte.readthedocs.io/}\\
    \midrule
    C9 & Support email for questions & \url{https://github.com/karnwatcharasupat/latte/issues}\\
    \bottomrule
\end{tabularx}
\else
\begin{tabular}{|l|p{6.5cm}|>{\raggedright\arraybackslash}p{9.5cm}|}
    \hline
    \textbf{Nr.} & \textbf{Code metadata description} &\\
    \hline
    C1 & Current code version & \publishedver \\
    \hline
    C2 & Permanent link to code/repository used for this code version & \url{https://github.com/karnwatcharasupat/latte}\\
    \hline
    C3  & Permanent link to Reproducible Capsule & \url{https://codeocean.com/capsule/3186531/tree}\\
    \hline
    C4 & Legal Code License   & MIT License \\
    \hline
    C5 & Code versioning system used & Git\\
    \hline
    C6 & Software code languages, tools, and services used & 
    Language: Python 3                  \newline
    CI/CD: pytest, CircleCI, CodeCov, CodeFactor            \\
    \hline
    C7 & Compilation requirements, operating environments \& dependencies & 
    Python $\ge$ 3.7, $<$ 3.10 \linebreak
    NumPy $\ge$ 1.18, Scikit-learn $\ge$ 1.0.0 \linebreak
    (optional)
    PyTorch $\ge$ 1.3.1, TorchMetrics $\ge$ 0.2.0 \linebreak
    (optional)
    TensorFlow $\ge$ 2.0 
    \\
    \hline
    C8 & If available Link to developer documentation/manual & \url{https://latte.readthedocs.io/}\\
    \hline
    C9 & Support email for questions & \url{https://github.com/karnwatcharasupat/latte/issues}\\
    \hline
\end{tabular}
\fi
\vskip0.5cm
\ifarxiv
\else
\end{@twocolumnfalse}
]
\fi

\clearpage
\newpage
\noindent

\sloppy
\section{Introduction}

Disentanglement learning and controllable generation are fast-growing fields within deep learning research, powered by the advances in deep generative networks, such as variational autoencoders (VAEs) \cite{Higgins2017-VAE:Framework} and generative adversarial networks (GANs) \cite{Chen2016InfoGAN:Nets}. Disentanglement learning is often used with encoder-decoder architectures to produce latent representations in the form of latent vectors or tensors in the bottleneck layer, such that each latent dimension has an approximately exclusive mapping to a semantic attribute of interest. These disentangled latent representations are particularly useful in the generative models that aim to produce samples with specific and controllable semantic attributes \cite{Pati2020Attribute-basedAuto-encoders, Tan2020MusicModelling}. 

With the growth of the fields comes the need for a reliable and consistent method of evaluation that allows for the comparison of different systems across a variety of metrics. Therefore, we introduce \Latte\footnote{\href{https://doi.org/10.5281/zenodo.5786402}{Software DOI: 10.5281/zenodo.5786402}} (for \textit{LATent Tensor Evaluation}), a cross-framework Python package for evaluation of latent-based generative models. Since successful latent-based controllable generation requires more than disentanglement~\cite{Pati2021IsGeneration}, \Latte also covers interpolatability metrics, in addition to disentanglement metrics.

Framework-agnostic evaluation tools are known to greatly facilitate and accelerate research development in a field. For example, in the field of audio source separation, \texttt{bss\_eval} \cite{Vincent2006PerformanceSeparation} and its successor \texttt{museval} \cite{Stoter2021Museval0.4.0} have greatly benefited the field and provided a standard benchmarking tool. Many studies on disentanglement learning have formally or informally relied on the \texttt{disentanglement\_lib}\footnote{\url{https://github.com/google-research/disentanglement_lib}} library for their evaluation. However, the library has not had a new release since 2019. Moreover, \texttt{disentanglement\_lib} was mainly created as a code base for reproducing the studies~\cite{Locatello2019ChallengingRepresentations, vanSteenkiste2019AreReasoning, Locatello2019OnRepresentations, Duan2019UnsupervisedLearning, Shu2020WeaklyGuarantees, Locatello2020DisentanglingLabels}, thus the metric implementations were written to fit the development codes rather than to cater to a wider range of applications, and are only available in TensorFlow. As a result, researchers working with PyTorch or other incompatible models often have to rely on their own re-implementations of the metrics -- an error-prone and inefficient approach that in the best case produces additional work, and in the worst case leads to inconsistencies in evaluation metric implementations. In addition, to the best of our knowledge, no comprehensive library for the evaluation of generative interpolatability currently exists.

The introduction of \Latte aims to address these shortcomings. By design, \Latte performs all metric calculations with NumPy-based computation to ensure cross-framework consistency. The modular design used in \Latte also ensures easy extensibility for supporting other frameworks beyond PyTorch and TensorFlow in the future.

\section{Software Description}

\Latte is a cross-framework Python package for the evaluation of disentanglement and controllability in latent-based generative models. \Latte supports on-the-fly metric calculation for disentanglement learning and controllable generation using both a standalone functional API, and modular APIs for the two major deep learning frameworks -- TensorFlow/Keras~\cite{Abadi2016TensorFlow:Learning} and PyTorch~\cite{Paszke2019PyTorch:Library}.

In order to maximize cross-framework compatibility and reproducibility, core functionalities of \Latte are developed with NumPy~\cite{Harris2020ArrayNumPy} and Scikit-learn~\cite{Pedregosa2011Scikit-learn:Python}, without any deep learning dependencies. These NumPy-based functionalities also serve as a standalone functional API, allowing the use of \Latte in post-hoc analyses without the need for specific deep learning dependencies like TensorFlow or PyTorch. 

For use with deep learning frameworks, we implemented a modular API for the metrics to allow easy usage within the respective framework. For use with TensorFlow/Keras, we implemented wrappers based on the Keras Metric API to convert the core NumPy-based functionalities to TensorFlow-compatible operations. With PyTorch, we implemented similar wrappers based on the TorchMetrics API~\cite{ThePyTorchLightningTeam2020TorchMetrics:Applications}, which allows easy integration with both PyTorch and the popular PyTorch Lightning~\cite{Falcon2019PyTorchLightning} frameworks. \Latte modular metrics can be used in distributed training using the respective built-in multi-node support in Keras and TorchMetrics. An example of using \textit{Latte} in modular mode with PyTorch is shown in \Cref{fig:torch}.

\begin{figure}[htb]
\begin{minted}[
    breaklines, 
    frame=single,
    framesep=10pt
]{python}
import latte
from latte.metrics.torch.disentanglement import MutualInformationGap
import torch

data = initialize_dataset()
model = initialize_model()

mig = MutualInformationGap(
    reg_dim=[0, 3, 7], 
    discrete=False
)

for x, attributes in range(data):
  xhat, z = model(x)
  mig.update(z=z, a=attributes)

mig_values = mig.compute()
\end{minted}
\caption{Example of using \Latte in modular mode with PyTorch via the TorchMetrics API. In this example, the data contains three continuous semantic attributes, which are respectively regularized by the latent specified by the argument \texttt{reg\_dim}. The \texttt{discrete=False} option specifies that the semantic attributes are continuous-valued.}
\label{fig:torch}
\end{figure}

\subsection{Deterministic metric calculation}
A number of metrics used in disentanglement learning and controllable generation are based on randomly-initialized regressors and classifiers \cite{Kumar2018VariationalObservations}. Moreover, practical calculation of probabilistic measures via Scikit-learn, such as mutual information and entropy, also requires random number generation. As \texttt{disentanglement\_lib} does not explicitly set a seed before metric calculation, identical inputs may result in different metric values. This particular detail is not commonly known amongst end-users who may not be aware of the implementation details. In \Latte, a random seed of \texttt{42} is set by default, but can be switched off by calling
\texttt{latte.seed(None)}.
This allows end-users to have deterministic metric calculation by default without having to know the implementation details of \Latte and its dependencies.

\subsection{Metric bundles}
In addition to individual metric functions and modules, \Latte provides metric bundles, which are special modules containing multiple metrics commonly used together, similar to \texttt{MetricCollection} in TorchMetrics. All metric submodules of a metric bundle are initialized together with a common set of settings, ensuring consistency and compatibility between the metrics within a bundle. Inputs of the update calls to a metric bundle are also automatically passed to the respective submodules, reducing the amount of codes needed for metric calculations. Custom bundles can also be created via the \texttt{MetricBundle} class in \Latte. An example of using a \Latte metric bundle with TensorFlow/Keras is shown in \Cref{fig:keras}.

\begin{figure}[htb]
\begin{minted}[
    breaklines, 
    frame=single,
    framesep=10pt
]{python}
import latte
from latte.metrics.keras.bundles import DependencyAwareMutualInformationBundle as DAMIBundle
import tensorflow as tf

data = initialize_dataset()
model = initialize_model()

bundle = DAMIBundle(
    reg_dim=[0, 3, 7], 
    discrete=False
)

for x, attributes in range(data):
  xhat, z = model(x)
  bundle.update_state(z=z, a=attributes)

metrics = bundle.result()

dmig_values = metrics['DMIG']
dlig_values = metrics['DLIG']
\end{minted}
\caption{Example of using a \Latte metric bundle in modular mode with TensorFlow via the Keras Metric API. The call signature is very similar to single-metric modules -- the main difference is that a metric bundle returns a dictionary of arrays instead of a single array. \texttt{DependencyAwareMutualInformationBundle} contains MIG, DMIG, XMIG, and DLIG. All individual metric submodules are automatically initialized with the same \texttt{reg\_dim} and \texttt{discrete} options.}
\label{fig:keras}
\end{figure}

\subsection{Testing and Deployment}

Automated testing of \Latte is performed via \texttt{pytest}. Continuous integration and deployment (CI/CD) is handled via CircleCI. Code coverage and code quality are respectively monitored via CodeCov and CodeFactor. \Latte releases are available on the Python Package Index (PyPI)\footnote{\url{https://pypi.org/project/latte-metrics/}} and can be easily installed via \texttt{pip install latte-metrics}.

\section{Supported Metrics}

Successful latent-based controllable generation requires more than disentanglement~\cite{Pati2021IsGeneration}. Evaluation of a controllable generative system generally falls into three categories: reconstruction, disentanglement, and interpolatability. \Latte currently focuses on disentanglement metrics and interpolatability metrics, since evaluation of reconstruction fidelity is domain-specific. Domain-specific reconstruction metrics for generative models may be added in future versions.

As of the current version (\publishedver), the following disentanglement metrics are supported: 
    Mutual Information Gap (MIG)~\cite{Chen2018IsolatingAutoencoders}, 
    Separate Attribute Predictability (SAP)~\cite{Kumar2018VariationalObservations}, 
    Modularity (Mod.)~\cite{Ridgeway2018LearningLoss},
    Dependency-aware Mutual Information Gap (DMIG)~\cite{Watcharasupat2021EvaluationAttributes},
    Dependency-blind Mutual Information Gap (XMIG), and
    Dependency-aware Latent Information Gap (DLIG)~\cite{Watcharasupat2021ControllableGeneration}.
The interpolatability metrics supported are: 
    Smoothness, and 
    Monotonicity~\cite{Watcharasupat2021ControllableGeneration}.
We briefly describe the supported metrics in the following sections.

\subsection{Disentanglement Metrics}

\subsubsection*{Mutual Information Gap} (MIG) evaluates the degree of disentanglement of a latent vector $\bm{z}\in\mathbb{R}^{D}$ with respect to a particular semantic attribute, $a_i\in\mathbb{R}$, by considering the gap in mutual information between the attribute and its most informative latent dimension and that between the attribute and its second-most informative latent dimension \cite{Chen2018IsolatingAutoencoders}. Mathematically, MIG is given by
\begin{equation}
    \text{MIG}(a_i, \bm{z})
    = \dfrac{\mathcal{I}(a_i, z_j)-\mathcal{I}(a_i, z_k)}{\mathcal{H}(a_i)},
\end{equation}
where $j=\argmax_d \mathcal{I}(a_i, z_d)$, $k=\argmax_{d \ne j} \mathcal{I}(a_i, z_d)$, $\mathcal{I}(\cdot, \cdot)$ is mutual information, and $\mathcal{H}(\cdot)$ is entropy.

\subsubsection*{Separate Attribute Predictability} (SAP) is similar in nature to MIG but, instead of mutual information, uses the coefficient of determination for continuous attributes and classification accuracy for discrete attributes to measure the extent of relationship between a latent dimension and an attribute \cite{Kumar2018VariationalObservations}. SAP is given by
\begin{equation}
    \text{SAP}(a_i, \bm{z})
    = \mathcal{S}(a_i, z_j)-\mathcal{S}(a_i, z_k),
\end{equation}
where $j=\argmax_d \mathcal{S}(a_i, z_d)$, $k=\argmax_{d \ne j} \mathcal{S}(a_i, z_d)$, and $\mathcal{S}(\cdot, \cdot)$ is either the coefficient of determination or classification accuracy.

\subsubsection*{Modularity} is a latent-centric measure of disentanglement \cite{Ridgeway2018LearningLoss} based on mutual information. Modularity measures the degree in which a latent dimension contains information about only one attribute, and is given by
\begin{equation}
    \text{Mod}(\{a_i\}, z_d)
    = 1 - \dfrac{\sum_{i \ne j}\left({\mathcal{I}(a_i, z_d)}/{\mathcal{I}(a_j, z_d)}\right)^2}{|\{a_i\}|-1},
\end{equation}
where $j=\argmax_{i}\mathcal{I}(a_i, z_d)$.

\subsubsection*{Dependency-aware Mutual Information Gap} (DMIG) is a dependency-aware version of MIG that accounts for attribute interdependence observed in real-world data \cite{Watcharasupat2021EvaluationAttributes}. Mathematically, DMIG is given by
\begin{equation}
    \text{DMIG}(a_i, \bm{z})
    = \dfrac{\mathcal{I}(a_i, z_j)-\mathcal{I}(a_i, z_k)}{\mathcal{H}(a_i|a_l)},
\end{equation}
where $j=\argmax_d \mathcal{I}(a_i, z_d)$, $k=\argmax_{d \ne j} \mathcal{I}(a_i, z_d)$, $\mathcal{H}(\cdot|\cdot)$ is conditional entropy, and $a_l$ is the attribute regularized by $z_k$. If $z_k$ is not regularizing any attribute, DMIG reduces to the usual MIG. DMIG compensates for the reduced maximum possible value of the numerator due to attribute interdependence.

\subsubsection*{Dependency-blind Mutual Information Gap} (XMIG) is a complementary metric to MIG and DMIG that measures the gap in mutual information with the subtrahend restricted to dimensions which do not regularize any attribute \cite{Watcharasupat2021ControllableGeneration}. XMIG is given by
\begin{equation}
    \text{XMIG}(a_i, \bm{z})
    = \dfrac{\mathcal{I}(a_i, z_j)-\mathcal{I}(a_i, z_k)}{\mathcal{H}(a_i)},
\end{equation}
where ${j=\argmax_d\mathcal{I}(a_i, z_d)}$, ${k=\argmax_{d \notin \mathcal{D}}\mathcal{I}(a_i, z_d)}$, and $\mathcal{D}$ is a set of latent indices which do not regularize any attribute. XMIG allows monitoring of latent disentanglement exclusively against attribute-unregularized latent dimensions.

\subsubsection*{Dependency-aware Latent Information Gap} (DLIG) is a latent-centric counterpart to DMIG \cite{Watcharasupat2021ControllableGeneration}. DLIG evaluates disentanglement of a set of semantic attributes $\{a_i\}$ with respect to a latent dimension $z_d$, such that
\begin{equation}
    \text{DLIG}(\{a_i\}, z_d)
    = \dfrac{\mathcal{I}(a_j, z_d)-\mathcal{I}(a_k, z_d)}{\mathcal{H}(a_j|a_k)},
\end{equation}
\mbox{where $j=\argmax_{i}\mathcal{I}(a_i, z_d)$, $k=\argmax_{i\ne j} \mathcal{I}(a_i, z_d)$.}

\subsection{Interpolatability Metrics}

The two interpolatability metrics currently supported are based on the concept of a pseudo-derivative called latent-induced attribute difference (LIAD), which is defined as
\begin{equation}
    \mathcal{D}_{i, d}(\bm{z}; \delta)
    = \dfrac{\mathcal{A}_i(\bm{z}+\delta \bm{e}_d) - \mathcal{A}_i(\bm{z})}{\delta}
\end{equation}
where $\mathcal{A}_i(\cdot)$ is the measurement of attribute $a_i$ from a sample generated from its latent vector argument, $d$ is the latent dimension regularizing $a_i$, $\delta>0$ is the latent step size, and $\bm{e}_d$ is the $d$th elementary vector~\cite{Watcharasupat2021ControllableGeneration}. Second-order LIAD is similarly defined by
\begin{equation}
    \mathcal{D}^{(2)}_{i, d}(\bm{z}; \delta)
    =\dfrac{{\mathcal{D}^{(1)}_i(\bm{z}+\delta \bm{e}_d) - \mathcal{D}^{(1)}_i(\bm{z})}}{\delta},
\end{equation}
where $\mathcal{D}^{(1)}\equiv\mathcal{D}$.

\subsubsection*{Smoothness} is a measure of how smoothly an attribute changes with respect to a change in the regularizing latent dimension \cite{Watcharasupat2021ControllableGeneration}. Smoothness of a latent vector $\bm{z}$ is based on the concept of second-order derivative, and is given by
\begin{equation}
    \text{Smoothness}_{i,d}(\bm{z}; \delta)
    = 1 - \dfrac{\mathcal{C}_{k\in\mathfrak{K}}\left[\mathcal{D}^{(2)}_{i,d}(\bm{\zeta}_k; \delta)\right]}{\delta^{-1}\mathcal{R}_{k\in\mathfrak{K}}\left[\mathcal{D}^{(1)}_{i,d}(\bm{\zeta}_k; \delta)\right]},
\end{equation}
where $\bm{\zeta}_k = \bm{z}+k\delta\bm{e}_d$, $\mathcal{C}_{k\in\mathfrak{K}}[\cdot]$ is the contraharmonic mean of its arguments over values of $k\in\mathfrak{K}$, and  $\mathcal{R}_{k\in\mathfrak{K}}[\cdot]$ is the range of its arguments over values of $k\in\mathfrak{K}$, and $\mathfrak{K}$ is the set of interpolating points used during evaluation.

\subsubsection*{Monotonicity} is a measure of how monotonic an attribute changes with respect to a change in the regularizing dimension \cite{Watcharasupat2021ControllableGeneration}. Monotonicity of a latent vector $\bm{z}$ is given by
\begin{equation}
    \text{Monotonicity}_{i,d}(\bm{z}; \delta, \epsilon) = \dfrac{\sum_{k\in\mathfrak{K}} I_k \cdot S_k}{\sum_{k\in\mathfrak{K}} I_k},
\end{equation}
where $S_k=\operatorname{sgn}\left(\mathcal{D}^{(1)}_{i,d}(\bm{z}+k\delta\bm{e}_d; \delta)\right)\in\{-1,0,1\}$, $I_k=\mathbb{I}\left[|\mathcal{D}^{(1)}_{i,d}(\bm{z}+k\delta\bm{e}_d; \delta)|>\epsilon\right]\in\{0, 1\}$, $\mathbb{I}[\cdot]$ is the Iverson bracket operator, and $\epsilon>0$ is a noise threshold for ignoring near-zero attribute changes.

\section{Software Impact and Future Work}

\Latte is released under the MIT License and welcomes community contribution to the package. The authors hope that the introduction of \Latte will reduce the amount of time spent on re-implementing evaluation metrics due to framework incompatibility, and provide a standardized and uniform framework for evaluation of controllable generative systems regardless of the deep learning framework of choice. 

The implementation of metrics such as Latent Density Ratio \cite{Pati2021IsGeneration} and Linearity \cite{Tan2020MusicModelling} is currently planned for future releases. Additional metrics under consideration include $\beta$-VAE Score \cite{Higgins2017-VAE:Framework}; FactorVAE Score \cite{Kim2018DisentanglingFactorising}; Explicitness \cite{Ridgeway2018LearningLoss}; Disentanglement, Completeness and Informativeness (DCI) Scores \cite{Eastwood2018ARepresentations}; Interventional Robustness Score (IRS) \cite{Suter2019RobustlyRobustness}; Consistency and Restrictiveness \cite{Shu2020WeaklyGuarantees}; and, Density and Coverage \cite{Naeem2020ReliableModels}. Wrapper supports for PyTorch Ignite \cite{Fomin2020High-levelPyTorch} and nnabla \cite{Hayakawa2021NeuralPerspectives} are also currently under consideration.

\section{Conclusion}

We introduce \Latte, a cross-framework Python package for evaluation of latent-based generative models. \Latte supports on-the-fly metric calculation for disentanglement learning and controllable generation using both standalone functional API based on NumPy, and modular APIs for both TensorFlow/Keras and PyTorch. \Latte eliminates the need for application-specific re-implementation of common metrics, allowing consistent and reproducible model evaluation regardless of the deep learning framework of choice. 

\processdelayedfloats

\section*{Declaration of competing interest}
The authors declare that they have no known competing financial interests or personal relationships that could have appeared to influence the work reported in this paper.

\section*{Acknowledgements}
J. Lee acknowledges the support from the CN Yang Scholars Programme, Nanyang Technological University, Singapore. Part of this work was done while K.~N.~Watcharasupat was also supported by the CN Yang Scholars Programme.

\noindent
\section*{References}
\renewcommand{\bibsection}{}

\bibliographystyle{IEEEtran}
\bibliography{references}

\end{document}